\documentclass[final,5p,times,twocolumn]{elsarticle}

\usepackage{amssymb}
\usepackage{amsmath}
\usepackage{tabularx}
\usepackage{url}
\usepackage{hyperref}

\makeatletter
\def\@date{}
\makeatother

\journal{}

\begin{document}

\begin{frontmatter}

\title{ACM-UNet: Adaptive Integration of CNNs and Mamba for Efficient Medical Image Segmentation}
\author[whut]{Jing Huang\corref{cor1}}
\author[whut]{Yongkang Zhao\fnref{fn1}}
\author[whut]{Yuhan Li\fnref{fn1}}
\author[cams]{Zhitao Dai\fnref{fn1}}
\author[whu]{Cheng Chen\fnref{fn1}}
\author[whut]{Qiying Lai\fnref{fn1}}
\cortext[cor1]{Corresponding author: Jing Huang (e-mail: huangjing@whut.edu.cn)}

\affiliation[whut]{
  organization={School of Computer Science and Artificial Intelligence, Wuhan University of Technology},
  city={Wuhan},
  postcode={430070},
  state={Hubei},
  country={China}
}

\affiliation[cams]{
  organization={National Cancer Center / National Clinical Research Center for Cancer / Cancer Hospital \& Shenzhen Hospital, Chinese Academy of Medical Sciences and Peking Union Medical College},
  city={Shenzhen},
  postcode={518116},
  state={Guangdong},
  country={China}
}

\affiliation[whu]{
  organization={Department of Radiation and Medical Oncology, Zhongnan Hospital of Wuhan University},
  city={Wuhan},
  postcode={430071},
  state={Hubei},
  country={China}
}

\begin{abstract}
The U-shaped encoder-decoder architecture with skip connections has become a prevailing paradigm in medical image segmentation due to its simplicity and effectiveness. While many recent works aim to improve this framework by designing more powerful encoders and decoders, employing advanced convolutional neural networks (CNNs) for local feature extraction, Transformers or state space models (SSMs) such as Mamba for global context modeling, or hybrid combinations of both, these methods often struggle to fully utilize pretrained vision backbones (e.g., ResNet, ViT, VMamba) due to structural mismatches.
To bridge this gap, we introduce ACM-UNet, a general-purpose segmentation framework that retains a simple UNet-like design while effectively incorporating pretrained CNNs and Mamba models
through a lightweight adapter mechanism. This adapter resolves architectural incompatibilities and enables the model to harness the complementary strengths of CNNs and SSMs—namely, fine-grained local detail extraction and long-range dependency modeling. Additionally, we propose a hierarchical multi-scale wavelet transform module in the decoder to enhance feature fusion and reconstruction fidelity.
Extensive experiments on the Synapse and ACDC benchmarks demonstrate that ACM-UNet achieves state-of-the-art performance while remaining computationally efficient. Notably, it reaches 85.12\% Dice Score and 13.89mm HD95 on the Synapse dataset with 17.93G FLOPs, showcasing its effectiveness and scalability. Code is available at: https://github.com/zyklcode/ACM-UNet.
\end{abstract}

\end{frontmatter}

\section{Introduction}
\label{Introduction}
Medical image segmentation plays a fundamental role in delineating anatomical structures and supporting downstream clinical decision-making. In recent years, rapid advances in deep learning have significantly driven progress in this domain. Since the introduction of the well-known U-Net~\cite{ronneberger2015u}, the U-shaped encoder-decoder architecture with skip connections has become the de facto standard for medical image segmentation~\cite{Zhou2018UNetAN, Chen2021TransUNetTM, Cao2021SwinUnetUP, Ruan2024VMUNetVM, Chen2024MSVMUNetMV}. Originally built upon convolutional neural networks (CNNs), this architecture offers strong capability for capturing fine-grained local features. Its skip connections effectively bridge low-level detail and high-level semantics, enabling accurate spatial localization.
Recently, diffusion models~\cite{shen2023advancing,shen2024imagpose,shen2025imagdressing} have emerged as a powerful generative paradigm in medical imaging tasks, offering superior performance in producing fine-grained anatomical details and handling data scarcity through conditional sampling.
Their~\cite{shen2025imaggarment,shen2025long} ability to model complex spatial distributions and incorporate multi-scale priors makes diffusion-based architectures a promising alternative or complementary backbone for segmentation frameworks.
However, with increasing structural complexity in medical images, pure CNN-based U-Net variants often fail to capture long-range dependencies, limiting their segmentation precision in complex scenarios.

To address this, many studies have extended the classic U-Net paradigm by enhancing encoder-decoder designs along three major directions. 
First, some works aim to strengthen convolutional expressiveness through architectural innovations. For example, the DeepLab series~\cite{chen2014semantic,chen2017deeplab,chen2017rethinking,chen2018encoder}, progressively enhances the Atrous Spatial Pyramid Pooling (ASPP) module to effectively capture multi-scale contextual information, while DCN~\cite{Dai2017DeformableCN} leverages deformable convolutions for adaptive receptive field modeling.
More recently, novel convolution operators such as DWConv~\cite{Howard2017MobileNetsEC}, CondConv~\cite{Yang2019CondConvCP}, Dynamic Convolution~\cite{Chen2019DynamicCA}, MixConv~\cite{Tan2019MixConvMD} and PinwheelConv~\cite{Yang2024PinwheelConv} have been proposed to further improve local feature extraction and flexibility. Despite their effectiveness in enhancing local features, these methods still lack the ability to model global context. 
Second, recent approaches directly integrate global modeling modules, such as Transformers~\cite{vaswani2017attention} or state space models (SSMs), like Mamba~\cite{Gu2023MambaLS}, into the encoder. For instance, Swin-UNet~\cite{Cao2021SwinUnetUP} exploits the Swin Transformer’s~\cite{Liu2021SwinTH} hierarchical self-attention for multi-scale global reasoning, while VM-UNet~\cite{Ruan2024VMUNetVM} incorporates the Visual State Space (VSS)~\cite{Liu2024VMambaVS} block to enable efficient long-range dependency modeling with linear complexity. 
Third, hybrid designs combine both CNNs and global modules to exploit their complementary strengths.For example, TransUNet~\cite{Chen2021TransUNetTM} extracts local features with CNNs and then refines global context using Transformers. TransFuse~\cite{Zhang2021TransFuseFT} proposes a dual-branch architecture that fuses CNN and Transformer features in a parallel and complementary manner, enabling effective integration of local detail and global semantics. Similarly, HC-Mamba~\cite{Xu2024HCMambaVM} introduces hybrid convolution-Mamba blocks to enhance both locality and contextual modeling in medical images.

\begin{figure*}[t]
\centering
\includegraphics[width=1.0\textwidth]{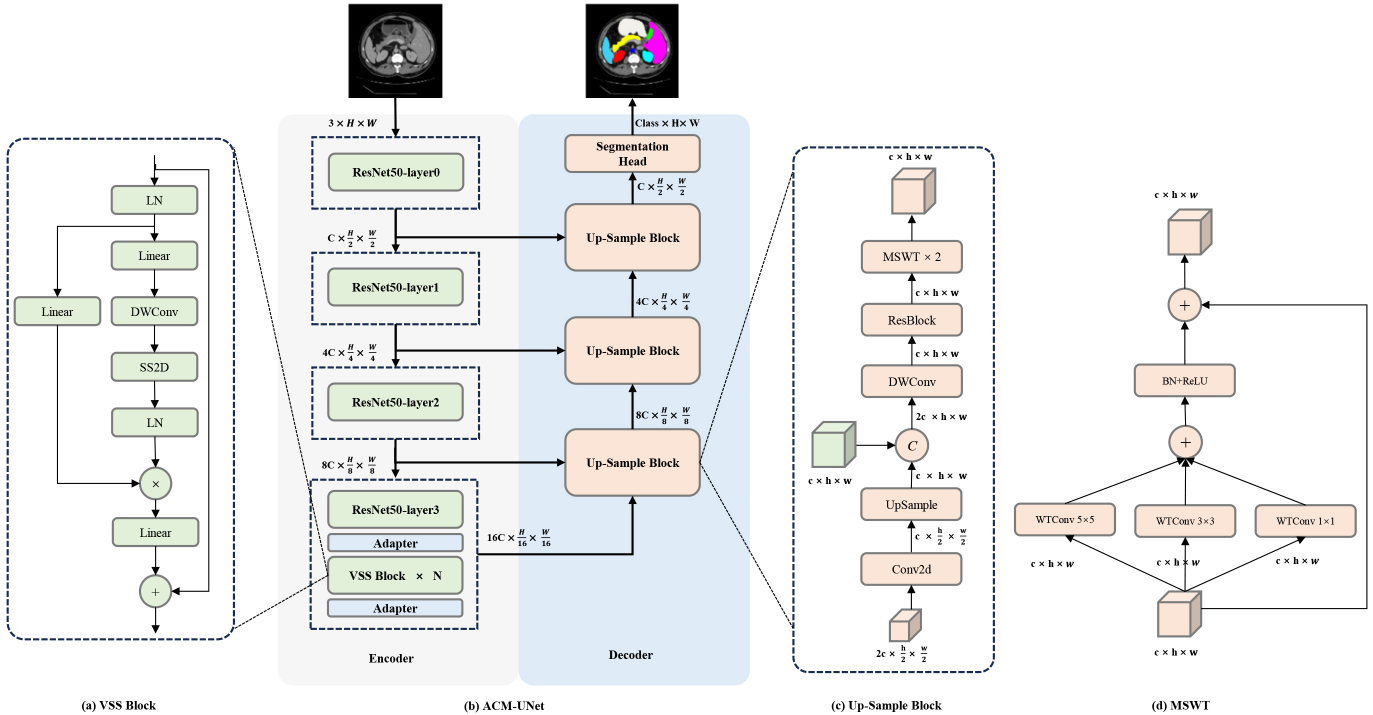}
\caption{The overall architecture of our proposed ACM-UNet. Left: (a) illustrates the structure of individual VSS block in its encoder, where LN and DWConv refer to layer normalization and depthwise separable convolution respectively, and SS2D is a 2D-selective-scan module. Middle: (b) details the encoder-decoder-style structure of ACM-UNet. Right: (c) depicts the inner details of each Up-Sample block in its decoder, where Conv2d achieves an ordinary $1 \times 1$ convolution, UpSample denotes 2 $\times$ up-sampling of spatial resolution via bilinear interpolation, and $C$ represents dimension concatenation. Right: (d) Structure of our multi-scale wavelet transform module.}\label{fig:model}
\end{figure*}

While these hybrid architectures have shown superior segmentation performance, they also introduce two key limitations. First, the increased architectural complexity often compromises efficiency, making them less suitable for real-world deployment. Second, most models adopt tightly coupled structures that prevent seamless integration of off-the-shelf pretrained backbones (e.g., ResNet~\cite{He2015DeepRL}, ViT~\cite{dosovitskiy2020image}, VMamba~\cite{Liu2024VMambaVS}). This limits the reusability of pretrained vision features, leading to increased training costs and reduced generalization. These issues pose significant challenges for clinical applications, where computational efficiency and model adaptability are crucial.

To overcome these limitations, we propose ACM-UNet, a simple yet general-purpose segmentation framework that adaptively integrates pretrained vision CNN and Mamba assets into a UNet-like architecture. 
As shown in Fig.~\ref{fig:model}(b), ACM-UNet follows the standard U-shaped encoder-decoder design. In the encoder, we incorporate well-established pretrained backbones, ResNet-50~\cite{He2015DeepRL} for local detail representation and VMamba~\cite{Liu2024VMambaVS} for modeling long-range semantics. To resolve structural discrepancies between CNN and Mamba blocks, a lightweight adapter is introduced, enabling smooth feature fusion and preserving computational efficiency.
To further enhance decoding, we propose a multi-scale wavelet transform module (MSWT), embedded at each upsampling stage of the decoder (Fig.~\ref{fig:model}(d)). This module refines hierarchical features and strengthens the fusion of spatial and contextual cues. 
Extensive experiments on two public benchmarks, Synapse and ACDC, validate the effectiveness of ACM-UNet. The model achieves superior segmentation accuracy and efficiency compared to state-of-the-art methods, while maintaining a simple and scalable design.

\textbf{Our contributions are summarized as follows:}
\begin{itemize}
    \item We propose \textbf{ACM-UNet}, a novel UNet-style segmentation framework that adaptively integrates pretrained ResNet and VMamba backbones via lightweight adapters. This design enables effective reuse of vision priors while maintaining model simplicity and generality.
    \item We develop a multi-scale wavelet transform module (MSWT) for the decoder, which enhances multi-level feature fusion and contributes to more accurate boundary preservation and segmentation performance.
    \item Extensive experiments on Synapse and ACDC datasets demonstrate that ACM-UNet outperforms existing state-of-the-art methods in segmentation accuracy while retaining a decent computational efficiency.
\end{itemize}

\section{Related Work}
\label{Related Work}
Our study employs two emerging techniques. In this section, we briefly introduce them.

\subsection{Visual State Space Block}
\label{Visual State Space Block}
\begin{figure*}[t]
    \centering
    \includegraphics[width=1.0\textwidth]{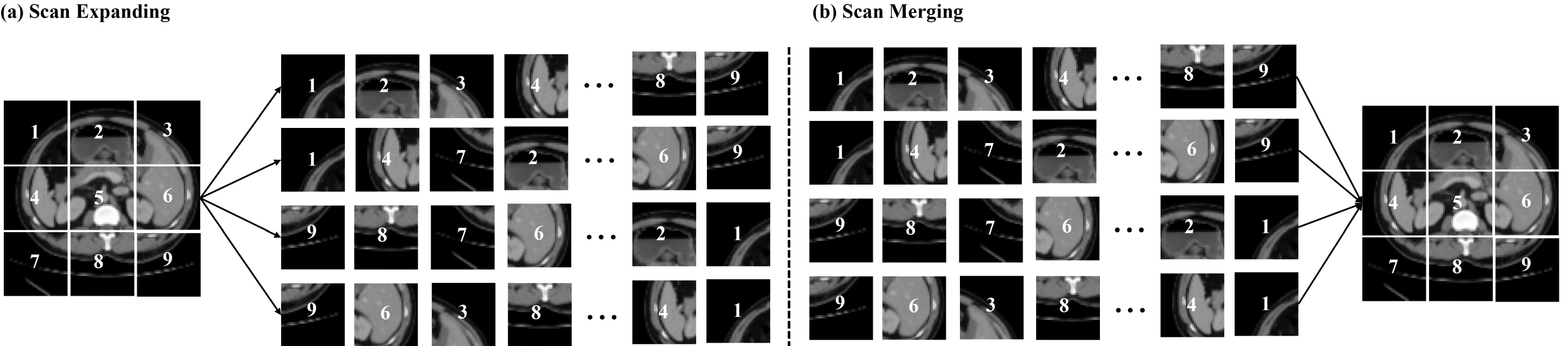}
    \caption{The scan expanding and scan merging process of medical images via 2D-selective-scan strategy. (a) presents the detail of scan expanding in our SS2D module; (b) depicts the corresponding scan merging scheme in our SS2D module.}
    \label{fig:ss2d}
\end{figure*}
State space models (SSMs) have evolved significantly in recent years, from early linear representations such as S4~\cite{Gu2021EfficientlyML} and its diagonal variant S4D~\cite{Gu2022OnTP}, to the more expressive and efficient Mamba~\cite{Gu2023MambaLS} architecture. These models offer a compelling alternative to attention-based Transformers, especially for long-range sequence modeling. To address the challenge of applying original Mamba~\cite{Gu2023MambaLS} to computer vision area, researchers have proposed several vision-specific extensions, including Vim~\cite{Zhu2024VisionME} and VMamba~\cite{Liu2024VMambaVS}. Vim~\cite{Zhu2024VisionME} introduces a visual backbone network with bidirectional Mamba blocks, demonstrating the potential of the Mamba architecture in visual tasks such as image classification.
Building on this, Yu et al. proposed a novel visual backbone network VMamba~\cite{Liu2024VMambaVS}, where visual state space block, i.e., the VSS block, is its fundamental component, similar to the role of individual residual layer in the ResNet~\cite{He2015DeepRL}. Figure \ref{fig:model}(a) presents the essential structure of the VSS block, in which SS2D is a called 2D-selective-scan module. The module is the key of realizing Mamba architecture in 2D imaging data, which consists of three successive steps: scan expanding, feature extraction via S6 block of Mamba, and scan merging. 

In this study, we employ the VSS blocks derived from prior well-pretrained VMamba to achieve long-range dependency modeling for global feature extraction within medical images. Figure \ref{fig:ss2d} illustrates the scan expanding and scan merging process of our medical image via four different scanning ways, such that contextual information of each image patch from different directions can be obtained. Specifically, as shown in Fig.\ref{fig:ss2d}(a), a medical image is firstly unfolded along four different directions (i.e., top-left to bottom-right, bottom-right to top-left, top-right to bottom-left, and bottom-left to top-right, respectively) into sequences during the scan expanding step. Next, the sequences are processed by the S6 block of Mamba that is essentially a specific implementation of state space model. Subsequently, as depicted in Fig.\ref{fig:ss2d}(b), the resulting sequences via S6 block from the four directions are merged, restoring the output image with same size as the input. 
\subsection{Wavelet Transform Convolution}
\label{Wavelet Transform Convolution}
Traditional convolutional kernels, operating at fixed scales, often struggle to adequately capture high-frequency details and low-frequency contours in signals. To address this, Wavelet Transform Convolution, i.e., WTConv~\cite{Finder2024WaveletCF}, is proposed recently, which combines the principles of wavelet transform and convolution, offering excellent capability of multi-scale feature extraction in signal and image processing. 

Specifically, the WTConv utilizes classic discrete wavelet transform to decompose the time-frequency characteristics of the input signal in parallel during the convolution process, enabling targeted processing of different frequency components. Here, the low\-frequency components are leveraged to learn global contextual information, while the high\-frequency components capture local details, thus strengthening feature representation capabilities. Additionally, by striking a balance between performance and computational efficiency, WTConv demonstrates strong potential for deployment in lightweight models as well. Thereby, in this work, we employ the WTConv to develop a specific pluggable module for our model's decoder to refine fused features while considering the efficiency of model.
\section{Method}
\label{Method}

\subsection{Overview of ACM-UNet}
\label{Overview of ACM-UNet}
To achieve competitive performance and efficiency while adhering to the intention of adopting a simple design strategy, we propose ACM-UNet in this paper. It still exploits the widely-accepted UNet-like architecture with encoder and decoder in medical image segmentation area. Figure \ref{fig:model} illustrates its network structure. Different from other UNet-like segmentation models, e.g., Swin-UNet~\cite{Cao2021SwinUnetUP}, TransUNet~\cite{Chen2021TransUNetTM} and HC-Mamba~\cite{Xu2024HCMambaVM}, our ACM-UNet prefers to simplicity and efficiency. 

In detail, the encoder of ACM-UNet is composed of four stages, the first three of which contains a module of ResNet50-layer respectively, and its final stage is a module sequence comprising an ResNet50-layer, two indispensable adapters and $N$ VSS blocks. Here, each ResNet50-layer (marked by a suffix $0\cdots3$ in Fig.\ref{fig:model}) is a stacked convolution-structural module representing the corresponding stage of original ResNet50~\cite{He2015DeepRL}, which consists of basic convolutions, ReLU activations, max-pooling operations, and residual connections. The VSS block comes from VMamba~\cite{Liu2024VMambaVS}, which is responsible for augmenting the above CNNs-based local features with global receptive fields through a specific selective scanning mechanism, e.g., SS2D. Moreover, to bridge the gap in the dimension and semantic of features between ResNet50-layer and VSS blocks, and guarantee the efficiency of our segmentation model as well, we introduce a lightweight adapter module, which is designed as a composite of common linear operations.  

Correspondingly, the decoder of ACM-UNet consists of three stages, each containing an Up-Sample block. Fig.\ref{fig:model}(c) details the network structure of individual Up-Sample block and the change of its inner feature dimensions. It receives representations from its previous stage,  and the output of  the corresponding stage in encoder associated by skip connection as well, and further concatenates them as the input of depthwise separable convolution (i.e., DWConv). Subsequently, after using an ordinary ResNet block - ResBlock, we intentionally introduce a multi-scale wavelet transform module, termed MSWT, to further refine prior fused representations. Finally, the segmentation result is obtained through a pithy segmentation head, which practically applies a 2D convolution operator. 
\subsection{Encoder with Concise and Generalized Design}
\label{Encoder with Concise and Generalized Design}
As shown in Fig.\ref{fig:model}(b), our encoder adopts a concise hybrid architecture, which hierarchically employs two widely-used general-purpose vision backbone networks, i.e., ResNet50~\cite{He2015DeepRL} and VMamba~\cite{Liu2024VMambaVS}. Here, its all modules relying on ResNet50 are responsible for extracting local features relative to medical details, while other modules based on VMamba are interested in capturing long-range dependencies of prior local features within medical images. Moreover, it also introduces a tailored module - adapter for dealing with the discrepancies of the above two kinds of modules in dimension and semantic of features. To reduce model cost, the adapter module pursues lightweight design strategy as well. Importantly, apart from the adapter, other modules in encoder, i.e., ResNet50-layer0$\ldots$3 and each VSS block, leverage those off-the-shelf assets pretrained on large-scale dataset (e.g., ImageNet-1K~\cite{Deng2009ImageNetAL}). The aim of such design is to facilitate training our encoder, while inheriting more fundamental vision knowledge from the generalized large-scale pretraining priors. As a result, the accuracy and generalization of model can be guaranteed.

In the encoder, the first three stages, i.e., ResNet50-layer0$\ldots$2, leverage the corresponding pretrained layers of original ResNet50, progressively extracting multi-scale features, denoted as $f_1^e$, $f_2^e$ and $f_3^e$ respectively. Concretely, our ResNet50-layer0 captures low-level local features $f_1^e  \in \mathbb{R}^{C \times \frac{H}{2} \times \frac{W}{2}}$ from input $X \in \mathbb{R}^{C\times H\times W}$ in Stage 1. The ResNet50-layer1 produces mid-level texture-related features $f_2^e \in \mathbb{R}^{4C \times \frac{H}{4} \times \frac{W}{4}}$ in Stage 2, and the ResNet50-layer2 offers semantic-related features $f_3^e \in \mathbb{R}^{8C \times \frac{H}{8} \times \frac{W}{8}}$ in Stage 3. Then, Stage 4 fuses the ResNet50-layer3 with two pretrained-well VSS blocks of VMamba via lightweight adapters, outputting the final representations of encoder $f_4^e \in \mathbb{R}^{16C \times \frac{H}{16} \times \frac{W}{16}}$. In the VSS blocks, we follow the 2D-selective-scan (SS2D) strategy of VMamba. Figure \ref{fig:ss2d} illustrates its application in our medical image segmentation task. 

Ultimately, by the above concise and generalized design, our encoder is characterized with the capability of efficiently capturing both local details and global contexts, which is critical to subsequent high-quality decoding for final accurate medical image segmentation.
\subsection{Multi-Scale Wavelet Transform Strategy}
\label{Multi-Scale Wavelet Transform Strategy}
Capturing robust and outstanding representations from medical images is critical to those semantic segmentation methods based on deep learning. Although both local details and global context within medical images are regarded in the above encoder via CNNs and Mamba, we still hope to further enhance the representation capability of our model. Therefore, we develop a novel module to accomplish multi-scale wavelet transform, termed MSWT, which can be independently incorporated into each stage of our decoding process for refining prior features. This strategy facilitates obtaining finer-grained representations for final precise segmentation. 

Figure \ref{fig:model}(d) presents the structure of our developed MSWT. In the MSWT, we parallelize three wavelet transform convolutions (i.e., WTConv)  of different kernel scales, each of which achieves the decomposition and reconstruction of representation signals in wavelet domain, and feature extraction by a specified 2D convolution as well. Specifically, for an input feature map $Z_{in}\in \mathbb{R}^{c\times h\times w}$, the MSWT first projects the spatial representation signal into the wavelet domain, and further decomposes it into the high-frequency and low-frequency components that represent local details and global context, respectively. Then, multiple 2D convolutional kernels $k_{i}$ ($i=0,1,2$ in MSWT) with different scales (e.g., $1\times1$, $3\times3$ or $5\times5$) are applied in these frequency components. Their convolutional results $x_{i}\in \mathbb{R}^{c\times h\times w}$ are integrated by addition, and subsequently handled by batch normalization (BN) and nonlinear activation (ReLU). Finally, an residual connection combines the fused features with the original input, producing the output of MSWT, i.e., refined representations $Z_{out}\in \mathbb{R}^{c\times h\times w}$. Formally, the entire process for MSWT is defined as follows:
\begin{equation}
\left\{
\begin{array}{l}
x_{i} = WTConv(Z_{in}, k_{i}) \\
Z_{out} = Z_{in} + ReLU(BN(\Sigma{x_{i}})) .
\end{array}
\right.
\end{equation}

This signal processing approach in dual-domain (i.e., frequency and spatial domains) enables effectively capturing  global structures and fine-grained details of medical images, while guaranteeing computational efficiency.
\subsection{Decoder with Feature Fusion Refinement}
\label{Decoder with Feature Fusion Refinement}
To enable obtaining efficiently high-resolution feature maps for ultimate precise semantic segmentation in medical images, we organize our decoder network in a progressively up-sampling and multi-layer feature-fusing manner based on previous encoding process. As illustrated in Fig.\ref{fig:model}(c), the Up-Sample block is its fundamental module, where the depthwise separable convolution (i.e., DWConv) is introduced to reduce parameter amount and improve computational efficiency. Importantly, in each Up-Sample block, we successively deploy two developed multi-scale wavelet transform (i.e., MSWT) modules to refine those fused features at current layer, further retaining more details critical to identifying organ margins. As a result, after three layers of Up-Sample blocks, the semantics of the representations in our decoder is definitely enriched. It is because both local details and global contexts within medical images are deliberately concerned during our multi-layer decoding process.   

Subsequently, those representations are fed into a lightweight segmentation head for final dense category prediction. The concise segmentation head first up-samples the feature maps to the target resolution via a basic up-sampling operator and a depthwise separable convolution, and then achieve the conversion from representation space to category distribution.
\section{Experiments and Results}
\label{Experiments and Results}

\subsection{Datasets}
We evaluate the performance of our proposed ACM-UNet on two modalities of medical image datasets, i.e., CT and MRI images. The two datasets are adopted widely in medical image segmentation algorithm assessment. They are the Synapse abdominal multi-organ segmentation dataset (Synapse) and the Automated Cardiac Diagnosis Challenge dataset (ACDC), respectively. 
\label{Datasets}
\subsubsection{\texorpdfstring{Synapse Dataset~\cite{landman2015miccai}}{Synapse Dataset}}
The dataset contains clinical CT images from 30 patients for abdominal multi-organ segmentation, totaling 3779 axial contrast-enhanced images. The images have an uniform resolution of $512 \times 512$ pixels, and offer accurate annotation of organ contours. Following TransUNet~\cite{Chen2021TransUNetTM}, we randomly split the dataset into 18 cases for training and 12 cases for testing, and focus on the segmentation of only 8 abdominal organs: the aorta, gallbladder, left kidney, right kidney, liver, pancreas, spleen, and stomach.

\subsubsection{\texorpdfstring{ACDC Dataset~\cite{Bernard2018DeepLT}}{ACDC Dataset}}
The dataset was from the automated cardiac diagnosis challenge, which collects cardiac MRI scan images from different patients. Each patient scan is annotated manually with ground truth for three sub-organs, i.e., left ventricle (LV), right ventricle (RV) and myocardium (Myo). Like TransUNet~\cite{Chen2021TransUNetTM}, we select 70 cases for training, 10 for validation, and 20 for testing.

\subsection{Implementation Details and Evaluation Metrics}
\label{Implementation Details and Evaluation Metrics}
\subsubsection{Implementation Details}

We implement ACM-UNet based on the PyTorch 2.1.1 framework, and conduct all experiments on a single NVIDIA GeForce RTX 3090 GPU. For each ResNet50-layer module in ACM-UNet, we leverage the corresponding layer of well-pretrained ResNet50 for initialization; for the VSS blocks in ACM-UNet, we utilize the pretrained weights from the ImageNet-1k dataset for initialization. During training, we employ extensive data augmentation strategies, including resizing input images to $224\times224$, horizontal and vertical flipping, random rotation, Gaussian noise, Gaussian blur, and contrast enhancement. The entire network is optimized using the AdamW optimizer, batch size of 32, and 300 epochs. The initial learning rate is set to 5e-4, and then is scaled down with the training process using the cosine annealing algorithm. Like MSVM-UNet~\cite{Chen2024MSVMUNetMV}, we set different weight decay values to reduce overfitting, where 1e-3 for the Synapse dataset and 1e-4 for the ACDC dataset. Meanwhile, we adopt a simple loss function for training, which is defined as a combination of Dice and cross-entropy (CE) loss functions as follows:
\begin{equation}
L_{total}=\alpha L_{Dice} + (1-\alpha)L_{CE},
\end{equation}
where $\alpha$ and $1-\alpha$ represent the weights of the Dice loss and cross-entropy loss, respectively.

\subsubsection{Evaluation Metrics}

We employ two commonly adopted metrics to assess the performance of ACM-UNet on the Synapse and ACDC datasets. They are the Dice Similarity Coefficient (DSC) and the 95$\%$ Hausdorff Distance (HD95), respectively. Mathematically, given a predicted segmentation mask $P$ and a ground truth segmentation mask $G$, the Dice Similarity Coefficient is defined as: 
\begin{equation}
DSC(P, G)=\frac{2\times |P\cap G|}{|P| + |G|},
\end{equation}
where $|P\cap G|$ represents the cardinality (number of elements) of the intersection between the predicted segmentation mask $P$ and the ground truth segmentation mask $G$, and $|P|$ and $|G|$ denote the cardinality of the predicted segmentation mask and the ground truth segmentation mask respectively. Next, let $A$ be the set of boundary points of the predicted segmented image and $B$ be the set of boundary points of the ground truth segmented image, and the Hausdorff Distance $H(A,B)$ is calculated as follows:
\begin{equation}
H(A,B)=\max\left\{\max_{a\in A}\min_{b\in B}|a - b|,\max_{b\in B}\min_{a\in A}|b - a|\right\}.
\end{equation}
To obtain HD95, we first calculate all the pairwise distances between points in $A$ and $B$ as described above, then sort these distances in ascending order, and finally select the value at the $95^{th}$ percentile of this sorted list.
\subsection{Performance comparisons with state-of-the-arts}
\label{Performance comparisons with state-of-the-arts}
To assess the performance of the proposed model, we conduct experiments on the above two datasets by comparing our ACM-UNet with some state-of-the-art (SOTA) models based on CNN, Transformer, and Mamba. 

\subsubsection{Results on the Synapse Dataset}
\begin{table*}[t] 
    \centering
    \resizebox{\textwidth}{!}{ 
    \begin{tabular}{| c | c | c | c | c | c | c | c | c | c | c | c |}
    \hline
\textbf{Methods} & \textbf{Year} & \textbf{$\mathbf{DSC}(\%)\!\boldsymbol{\uparrow}$} & \textbf{$\mathbf{HD}95\,(\mathbf{mm})\!\boldsymbol{\downarrow}$} & \textbf{Aorta} & \textbf{Gallbladder} & \textbf{Kidney(L)} & \textbf{Kidney(R)} & \textbf{Liver} & \textbf{Pancreas} & \textbf{Spleen} & \textbf{Stomach} \\
        \hline
UNet~\cite{ronneberger2015u}&2015&76.85&39.70&85.66&53.24&81.13&71.60&92.69&56.81&87.46&69.93\\
        Unet++~\cite{Zhou2018UNetAN}&2018&76.91&36.93&88.19&68.89&81.76&75.27&93.01&58.20&83.44&70.52\\
        Att-UNet~\cite{Rahman2023MedicalIS}&2018&71.70&34.47&82.61&61.94&76.07&70.42&87.54&46.70&80.67&67.66\\
        TransUNet~\cite{Chen2021TransUNetTM}&2021&76.76&44.31&86.71&58.97&83.33&77.95&94.13&53.60&84.00&75.38\\
        Swin-UNet~\cite{Cao2021SwinUnetUP}&2021&79.13&21.55&85.47&66.53&83.28&79.61&94.29&56.58&90.66&76.60\\
        LeVit-Unet-384~\cite{xu2023levit}&2021&78.53&16.84&87.33&62.23&84.61&80.25&93.11&59.07&88.86&72.76\\
        DAE-Former~\cite{azad2023dae}&2022&82.63&16.39&87.84&71.65&87.66&82.39&95.08&63.93&91.82& 80.77\\
        ScaleFormer~\cite{huang2022scaleformer}&2022&82.86&16.81&88.73&\textbf{74.97}&86.36&83.31&95.12&64.85&89.40&80.14\\
        PVT-CASCADE~\cite{Rahman2023MedicalIS}&2023&81.06&20.23&83.01&70.59&82.23&80.37&94.08&64.43&90.10&83.69\\
        TransCASCADE~\cite{Rahman2023MedicalIS} &2023&82.68&17.34&86.63&68.48&87.66&84.56&94.43&65.33&90.79&83.52\\
        2D D-LKA Net~\cite{Azad2023BeyondSD}&2024&84.27&20.04&88.34&73.79&\underline{88.38}&\textbf{84.92}&94.88&67.71&91.22&\underline{84.94}\\
        MERIT-GCASCADE~\cite{Rahman2023GCASCADEEC}&2024&84.54&\textbf{10.38}&88.05&74.81&88.01&\underline{84.83}&95.38&69.73&91.92&83.63\\
        PVT-EMCAD-B2~\cite{Wang2021PVTVI}&2024&83.63&15.68&88.14&68.87&88.08&84.10&95.26&68.51&\underline{92.17}&83.92\\
        VM-UNet~\cite{Ruan2024VMUNetVM}&2024&82.38&16.22&87.00&69.37&85.52&82.25&94.10&65.77&91.54&83.51\\
        HC-Mamba~\cite{Xu2024HCMambaVM}&2024&79.58&26.34&\underline{89.93}&67.65&84.57&78.27&95.38&52.08&89.49&79.84\\

        Swin-UMamba~\cite{Liu2024SwinUMambaMU}&2024&82.26&19.51&86.32&70.77&83.66&81.60&95.23&69.36&89.95&81.14\\
        MSVM-UNet~\cite{Chen2024MSVMUNetMV}&2024&\underline{85.00}&14.75&88.73&\underline{74.90}&85.62&84.47&\textbf{95.74}&\textbf{71.53}&\textbf{92.52}&\textbf{86.51}\\
        MixFormer~\cite{Liu2025MixFormerAM}&2025&82.64&\underline{12.67}&87.36&71.53&86.22&83.19&95.23&66.82&89.98&80.77\\
        \hline
        ACM-UNet(ours)&2025&\textbf{85.12}&13.89&\textbf{90.04}&72.10&\textbf{90.48}&84.81&\underline{95.64}&\underline{70.96}&92.10&84.82\\
        \hline
   \end{tabular}
   }
    \caption{Comparative Experimental Results of Our ACM-UNet and SOTA Models on the Synapse Dataset. Here, Bold Black Data indicates the Best Result, and Underlined Black Data Denotes is the Second-Best Result.}
    \label{tab:synapse-compare}
\end{table*}
Table \ref{tab:synapse-compare} presents our performance comparison with various types of methods on the Synapse dataset. The proposed ACM-UNet achieves the best average DSC of 85.12\% and HD95 of 13.89mm. In detail, compared to CNN-based methods (e.g., 2D D-LKA Net \cite{Azad2023BeyondSD}), our ACM-UNet improves DSC and HD95 by 0.85\% and 6.15mm, respectively; compared to Transformer-based methods (e.g., PVT-EMCAD-B2 \cite{Rahman2023GCASCADEEC}), ACM-UNet  achieves improvements of 1.49\% and 1.79mm, respectively; compared to Mamba-based methods (e.g., MSVM-UNet \cite{Chen2024MSVMUNetMV}), it achieves improvements of 0.12\% and 0.86mm, respectively. Particularly, for the challenging segmentation targets (such as aorta, left kidney), ACM-UNet surpasses the 90\% DSC threshold, achieving 90.04\% and 90.48\%, respectively, which are 1.27\% and 2.10\% higher than the previous best results. This thanks to the specialty of ACM-UNet in marrying the local feature extraction capability of CNNs with the long-range dependency modeling capability of Mamba. Moreover, due to the introduction of multi-scale wavelet transform (MSWT) module, those features are further enhanced. As a result, the complex structures of the aorta and left kidney are effectively captured, further achieving significant performance improvements in segmenting the two difficult targets. 

\subsubsection{Results on the ACDC Dataset}
\begin{table}[t] 
    \centering
    \resizebox{0.5\textwidth}{!}{ 
    \begin{tabular}{| c | c | c | c | c | c |}
        \hline
        \textbf{Methods} & \textbf{Year} & \textbf{$\mathbf{DSC(\%)\!\uparrow}$} & \textbf{RV} & \textbf{Myo} & \textbf{LV} \\
        \hline
        R50 UNet~\cite{Chen2021TransUNetTM}&2021&87.55&87.10&80.63&94.92\\
        R50 Att-UNet~\cite{Chen2021TransUNetTM}&2021&86.75&87.58&79.20&93.47\\
        TransUNet~\cite{Chen2021TransUNetTM}&2021&89.71&88.86&84.53&95.73\\
        Swin-Unet~\cite{Cao2021SwinUnetUP}&2021&90.00&88.55&85.62&95.83\\
        ScaleFormer~\cite{huang2022scaleformer}&2022&90.17&87.33&88.16&95.04\\
        PVT-CASCADE~\cite{Rahman2023MedicalIS}&2023&91.46&88.90&89.97&95.50\\
    TransCASCADE~\cite{Rahman2023MedicalIS}&2023&91.63&89.14&\underline{90.25}&95.50\\
        MERIT-GCASCADE~\cite{Rahman2023GCASCADEEC}&2024&92.23&90.64&89.96&96.08\\
        PVT-EMCAD-B2~\cite{Rahman2023GCASCADEEC}&2024&92.12&90.65&89.68&96.02\\
        VM-Unet~\cite{Ruan2024VMUNetVM}&2024&92.24&90.74&89.93&96.03\\
        Swin-Umamba~\cite{Liu2024SwinUMambaMU}&2024&92.14&\underline{90.90}&89.80&95.72\\
        MSVM-UNet~\cite{Chen2024MSVMUNetMV}&2024&\textbf{92.58}&\textbf{91.00}&\textbf{90.35}&\textbf{96.39}\\
        MixFormer~\cite{Liu2025MixFormerAM}&2025&91.01&89.02&88.46&95.55\\
        \hline
        ACM-UNet(ours)&2025&\underline{92.29}&90.61&89.94&\underline{96.31}\\
        \hline
   \end{tabular}
   }
    \caption {Comparative Experimental Results of Our ACM-UNet and SOTA Models on the ACDC Dataset. Here, Bold Black Data Indicates the Best Result, and Underlined Black Data Denotes is the Second-Best Result.}
    \label{tab:acdc-compare}
\end{table}
Table \ref{tab:acdc-compare} shows the performance  comparative results  of our ACM-UNet and somel representative advanced methods on the ACDC dataset. The experimental results demonstrate that ACM-UNet performs exceptionally well in the segmentation task of left ventricle (LV), achieving the DSC of 96.31\%. Although the result is slightly lower than the 96.39\% of MSVM-UNet \cite{Chen2024MSVMUNetMV}, ACM-UNet still significantly outperforms other competing methods. Overall, ACM-UNet achieves the impressive average DSC of 92.29\%, showcasing its robust capability in cardiac structure segmentation. Furthermore, for the segmentation tasks of right ventricle (RV) and myocardium (Myo), ACM-UNet also achieves the wonderful DSC results of 90.61\% and 89.94\%, respectively, which are slightly lower than the best-performing MSVM-UNet \cite{Chen2024MSVMUNetMV}. However, considering its computational efficiency, ACM-UNet remains highly competitive.

Meanwhile, the experimental results also verify the generalizability of our ACM-UNet like other excellent methods (e.g., aforementioned MSVM-UNet), since it performs well on different modalities of medical image data (CT and MRI).
\subsection{Model Complexity and Efficiency Analysis}
\label{Model Complexity and Efficiency Analysis}
\begin{table}[ht] 
    \centering
    \resizebox{0.5\textwidth}{!}{ 
    \begin{tabular}{| c | c | c | c |}
        \hline
        Methods & Year & \#Params(M)& \#FLOPs(G) \\
        \hline
        TransUNet \cite{Chen2021TransUNetTM} &2021& 96.07 & 88.91 \\
        Swin-UNet \cite{Cao2021SwinUnetUP} &2021& \underline{27.17} & \textbf{6.16} \\
        LeVit-Unet-384~\cite{xu2023levit}&2021&52.17&25.55\\
        DAE-Former~\cite{azad2023dae}&2022&48.07&27.89\\
        ScaleFormer~\cite{huang2022scaleformer}&2022&111.6&48.93\\
        2D D-LKA Net \cite{Azad2023BeyondSD} &2024& 101.64 & 19.92 \\
        MSVM-UNet \cite{Chen2024MSVMUNetMV} &2024& 35.93 & \underline{15.53} \\
        MixFormer~\cite{Liu2025MixFormerAM}&2025&123.35&108.32\\
        \hline
        ACM-UNet(ours) &2025& \textbf{16.48} & 17.93 \\
        \hline
    \end{tabular}
    }
    \caption{Comparison of Model Complexity and Efficiency Between ACM-UNet and Other SOTA Models. Here, Bold Black Data indicates the Best Result, and Underlined Black Data Denotes is the Second-Best Result.} 
    \label{tab:complex-compare} 
\end{table}
We investigate model complexity and efficiency by comparing the parameter scale and computational cost of ACM-UNet with several classic SOTA methods. Likewise, the methods are based on CNNs, Transformer and Mamba, respectively. Here, we use number of parameters to stand for the parameter scale of model, and employ FLOPs (i.e., floating point operations per second) as a metric to evaluate the computational cost of each deep learning model. Additionally, the number of parameters and FLOPs of the reported models are calculated using a popular python package - calflop with an input size of $224\times224\times3$.

As shown in Table~\ref{tab:complex-compare}, ACM-UNet presents outstanding performance in parameter scale. Despite its computational cost is not the best, it still achieves competitive results. Specifically, the parameter number of ACM-UNet is only 16.48M, which is significantly lower than ones of the TransUNet (96.07M) and the 2D D-LKA Net (101.6M), and also lower than the parameter scales of Swin-UNet (27.17M) and MSVM-UNet (35.93M). Meanwhile, for the computational cost, ACM-UNet achieves the FLOPs of 17.93G, which is considerably lower than TransUNet (88.91G) and MixFormer (108.32G), and not much different from 2D D-LKA Net (19.92G) and MSVM-UNet (15.53G). The results also demonstrate the efficiency of our ACM-UNet, which is critical for the clinic application of a segmentation model, particularly in real-time or resource-constrained settings.

\subsection{Qualitative Analysis}
\label{Qualitative Analysis}
\begin{figure*}[ht]
    \centering
    \includegraphics[width=0.8\textwidth]{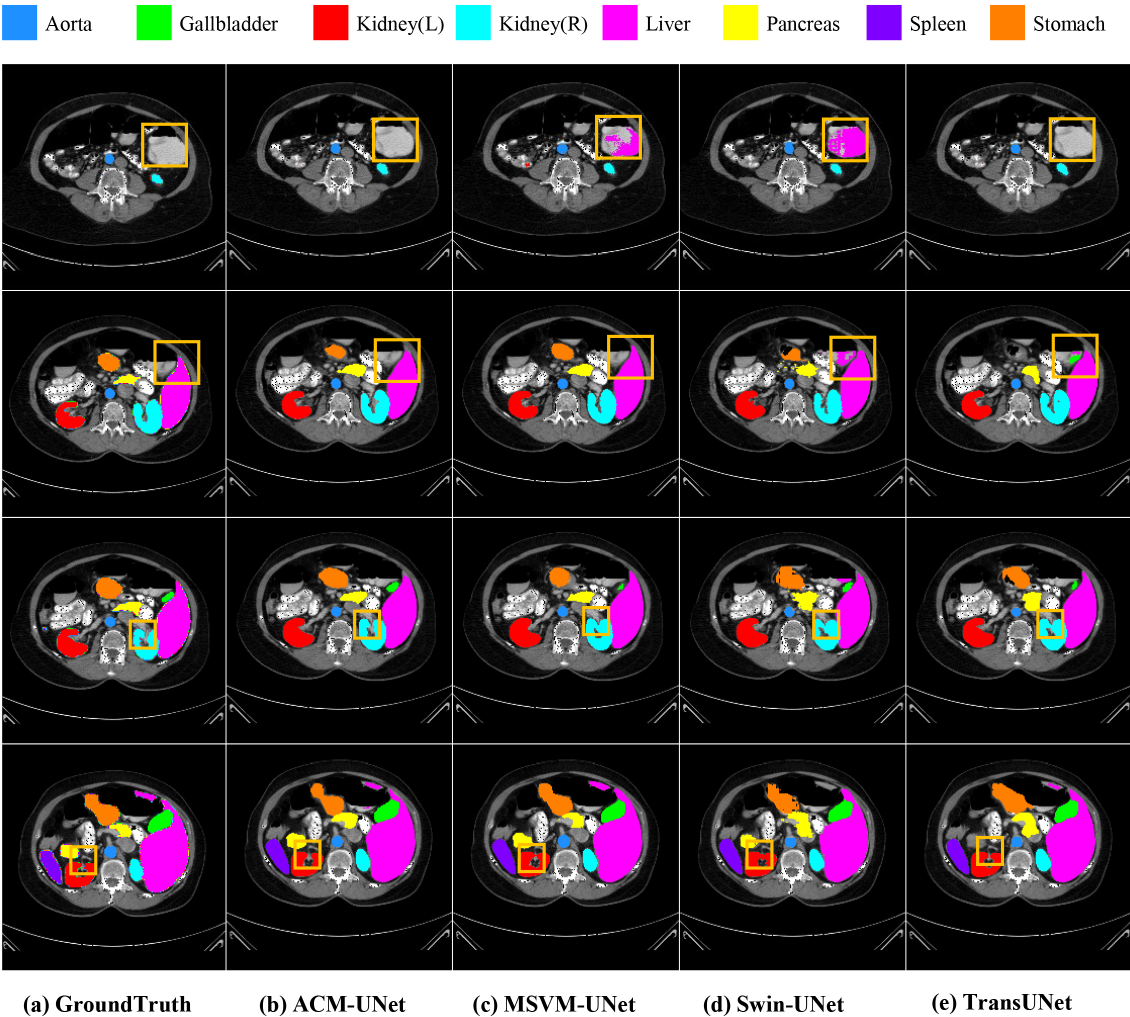}
    \caption{Visual comparison of different methods on the Synapse multi-organ dataset. The first column represents the ground truth, and the following columns represent the segmentation predictions of ACM-UNet, MSVM-UNet, Swin-UNet, and TransUNet, respectively. It can be observed that our ACM-UNet is of superiority in those organ regions circles by orange rectangles. }
    \label{fig:visual}
\end{figure*}
To conduct qualitative analysis, we carry out a 2D visual comparison of our ACM-UNet with several SOTAs on the Synapse multi-organ dataset. Figure~\ref{fig:visual} visualizes the comparison results. It can be observed that our model presents better segmentation results on the organs, especially their boundaries are delineated quite appropriately, such as the segmentation of the left kidney and right kidney in the last two rows. The reason is that our ACM-UNet better leverages the well-pretrained convolution network and is more sensitive to the vision details. Moreover, compared to other methods, our model avoids the over-segmentation of large organs to a certain extent (e.g., the segmentation of background in the first row and the liver in the second row). This is due to the introduction of the well-pretrained VSS blocks via tailored adapter, which makes up for the lack of global dependency modeling of the previous convolution network, thus producing more discriminative feature representations and obtaining better segmentation results.
\subsection{Ablation Studies}
\label{Ablation Studie}
Like numerous SOTA methods, our proposed ACM-UNet still follows the popular architecture of medical image segmentation, i.e., UNet-like encoder-decoder structure. However, different from those methods based on CNNs, Transformer or Mamba, except for chasing concise design scheme to carry out the UNet-like model architecture, we intentionally introduce two key components to our ACM-UNet: the pretrained VSS block for encoder and the tailored MSWT module for decoder. To verify the significancy of the two components, we conduct a comprehensive ablation study of ACM-UNet on the Synapse dataset. 
\subsubsection{Impact of Visual State Space Block and Multi-Scale Wavelet Transform Module on ACM-UNet Performance}

\begin{table*}[t] 
    \centering
    \resizebox{0.7\textwidth}{!}{ 
    \begin{tabular}{| c | c | c | c | c | c |}
        \hline
       \textbf{MSWT modules} & \textbf{VSS blocks} & \textbf{DSC(\%)$\boldsymbol{\uparrow}$} & \textbf{HD95(mm)$\boldsymbol{\downarrow}$} & \textbf{\#FLOPs(G)} & \textbf{\#Params(M)} \\
       \hline
                      &               & 83.52 & 25.95 & 10.46 & 11.04\\
        $\checkmark$ &               & 84.68 & 13.80 & 16.71 & 13.35\\
        & $\checkmark$               & 84.86 & 11.14 & 11.68 & 14.17\\
        $\checkmark$ &$\checkmark$    & 85.12 & 13.89 & 17.93 & 16.48\\
       \hline
   \end{tabular}
   }
    \caption{Ablation Study on the Impact of MSWT Modules and VSS Blocks on ACM-UNet Performance on the Synapse Dataset}
    \label{tab:all-ablation}
\end{table*}
We arrange a series of experiments to assess the effect w/o the two components in ACM-UNet. Table \ref{tab:all-ablation} presents the results of ablation study on the impact of VSS block and MSWT module on ACM-UNet performance. The experimental results demonstrate that the introduction of the two components significantly improves the performance of segmentation model. In detail, when ACM-UNet deploys MSWT modules alone, the DSC increases from 83.52\% to 84.68\%, and the HD95 significantly decreases from 25.95mm to 13.80mm, indicating that our multi-scale feature extraction strategy effectively enhances the model’s ability to capture complex structures and boundary regions of organs; when VSS blocks are employed alone, the DSC has further increased to 84.86\% while the HD95 decreased to 11.14mm, demonstrating that our global context modeling scheme further optimizes the segmentation effect, especially in localizing boundary regions of organs; when both MSWT modules and VSS blocks are applied in ACM-UNet, the entire model achieves its best performance, with the DSC of 85.12\% and the HD95 of 13.89mm. Although the FLOPs representing computational cost increases from 10.46G to 17.93G and the number of parameters increases from 11.04M to 16.48M, the performances (DSC and HD95) of ACM-UNet are clearly bettered, especially in handling complex anatomical structures such as the left kidney and aorta.

Therefore, the importance of our MSWT Modules and VSS blocks is validated for enhancing the segmentation capability of ACM-UNet. It also demonstrates that our combination of multi-scale feature extraction and global context modeling can effectively address key challenges in medical image segmentation.
\subsubsection{Impact of the Number of VSS Blocks on ACM-UNet Performance}

\begin{table*}[t]
    \centering
    \resizebox{0.6\textwidth}{!}{ 
    \begin{tabular}{| c | c | c | c | c |}
    \hline
    \textbf{\#VSS blocks} & \textbf{DSC (\%) $\uparrow$} & \textbf{HD95 (mm) $\downarrow$} & \textbf{\#FLOPs (G)} & \textbf{\#Params (M)} \\
    \hline
        8 & 84.86 & 18.92 & 21.60 & 25.86 \\
        4 & 84.43 & 15.58 & 19.16 & 19.61 \\
        2 & 85.12 & 13.89 & 17.93 & 16.48 \\
    \hline
    \end{tabular}
    }
    \caption{Ablation Study on the Impact of the Number of VSS Blocks on ACM-UNet Performance on the Synapse Dataset.}
    \label{tab:vss-ablation}
\end{table*}

The capability of the VSS block in capturing global vision contexts has been demonstrated in VMamba \cite{Liu2024VMambaVS}, and its important role in our ACM-UNet is also verified in the above experiments. Accordingly, in this subsection, we investigate the impact of the number of VSS blocks within ACM-UNet on the model performance through ablation experiments. 

Table \ref{tab:vss-ablation} shows the experimental results, which clearly illustrate the significant influence of different numbers of VSS blocks on the model performance and efficiency. It is observed that as the number of VSS blocks increases, the parameter scale of our model increases significantly, but the model performance does not exhibit a monotonic improvement trend. Specifically, given two VSS blocks, ACM-UNet achieves the highest DSC of 85.12\% and the lowest HD95 of 13.89mm, while maintaining the relatively low computational complexity (FLOPs of 17.93G) and number of parameters (16.48M). In contrast, given four VSS blocks, the DSC slightly decreases to 84.43\%, and the HD95 increases to 15.58mm. When deploying eight VSS blocks in ACM-UNet, the DSC further decreases to 84.86\%, whereas the HD95 increases to 18.92mm, and the parameter scale significantly increases to 25.86M. In brief, the results indicate that an excessive number of VSS blocks leads to an imbalance between global and local features, thereby limiting the performance improvement of our segmentation model. It follows that an appropriate number of VSS blocks can ensure satisfactory performance while enabling the modest tradeoff of local feature extraction and long-range dependency modeling.

It can be also seen that two VSS blocks is a decent configuration for our ACM-UNet, which can offer an impressive segmentation performance and efficiency, i.e., the DSC of 85.12\% and the HD95 of 13.89mm with 17.93G FLOPs and 16.48M parameter number scale
on the Synapse dataset. Importantly, the setting is competent for those medical image segmentation tasks with real-time needs or limited computational resources.
\section{Conclusion}
\label{Conclusion}
To facilitate offering a simple and general model for medical image segmentation tasks, we propose the ACM-UNet, a pretraining-guided adaptive CNN-Mamba UNet. ACM-UNet follows the classic U-shaped encoder-decoder-style architecture, yet adheres to the intention of concise design. Specifically, in the encoder of our UNet-like structure, ACM-UNet integrates straightly the off-the-shelf assets of two outstanding general-purpose feature extractors via a lightweight adapter to capture local details and global representations within medical images. Meanwhile, to enhance the decoding capability of ACM-UNet, we also deliberately design a multi-scale wavelet transform module, which is introduced hierarchically into our decoder for feature fusion refinement. Experimental results demonstrate that compared with recent SOTA models, our proposed ACM-UNet achieves competitive performance and computational efficiency while maintaining less parameter scale. Our future work will focus on further optimizing model complexity and boosting its performance and efficiency to enable broader application in real-time or resource-constrained clinical scenarios.

\bibliographystyle{elsarticle-num} 
\bibliography{references}

\begin{thebibliography}{10}
\expandafter\ifx\csname url\endcsname\relax
  \def\url#1{\texttt{#1}}\fi
\expandafter\ifx\csname urlprefix\endcsname\relax\def\urlprefix{URL }\fi
\expandafter\ifx\csname href\endcsname\relax
  \def\href#1#2{#2} \def\path#1{#1}\fi

\bibitem{ronneberger2015u}
O.~Ronneberger, P.~Fischer, T.~Brox, U-net: Convolutional networks for biomedical image segmentation, in: Medical image computing and computer-assisted intervention--MICCAI 2015: 18th international conference, Munich, Germany, October 5-9, 2015, proceedings, part III 18, Springer, 2015, pp. 234--241.

\bibitem{Zhou2018UNetAN}
Z.~Zhou, M.~M. Rahman~Siddiquee, N.~Tajbakhsh, J.~Liang, Unet++: A nested u-net architecture for medical image segmentation, in: Deep learning in medical image analysis and multimodal learning for clinical decision support: 4th international workshop, DLMIA 2018, and 8th international workshop, ML-CDS 2018, held in conjunction with MICCAI 2018, Granada, Spain, September 20, 2018, proceedings 4, Springer, 2018, pp. 3--11.

\bibitem{Chen2021TransUNetTM}
J.~Chen, Y.~Lu, Q.~Yu, X.~Luo, E.~Adeli, Y.~Wang, L.~Lu, A.~L. Yuille, Y.~Zhou, Transunet: Transformers make strong encoders for medical image segmentation, ArXiv abs/2102.04306 (2021).

\bibitem{Cao2021SwinUnetUP}
H.~Cao, Y.~Wang, J.~Chen, D.~Jiang, X.~Zhang, Q.~Tian, M.~Wang, Swin-unet: Unet-like pure transformer for medical image segmentation, in: European conference on computer vision, Springer, 2022, pp. 205--218.

\bibitem{Ruan2024VMUNetVM}
J.~Ruan, S.~Xiang, Vm-unet: Vision mamba unet for medical image segmentation, ArXiv abs/2402.02491 (2024).

\bibitem{Chen2024MSVMUNetMV}
C.~Chen, L.~Yu, S.~Min, S.~Wang, Msvm-unet: Multi-scale vision mamba unet for medical image segmentation, 2024 IEEE International Conference on Bioinformatics and Biomedicine (BIBM) (2024) 3111--3114.

\bibitem{shen2023advancing}
F.~Shen, H.~Ye, J.~Zhang, C.~Wang, X.~Han, W.~Yang, Advancing pose-guided image synthesis with progressive conditional diffusion models, arXiv preprint arXiv:2310.06313 (2023).

\bibitem{shen2024imagpose}
F.~Shen, J.~Tang, Imagpose: A unified conditional framework for pose-guided person generation, Advances in neural information processing systems 37 (2024) 6246--6266.

\bibitem{shen2025imagdressing}
F.~Shen, X.~Jiang, X.~He, H.~Ye, C.~Wang, X.~Du, Z.~Li, J.~Tang, Imagdressing-v1: Customizable virtual dressing, in: Proceedings of the AAAI Conference on Artificial Intelligence, Vol.~39, 2025, pp. 6795--6804.

\bibitem{shen2025imaggarment}
F.~Shen, J.~Yu, C.~Wang, X.~Jiang, X.~Du, J.~Tang, Imaggarment-1: Fine-grained garment generation for controllable fashion design, arXiv preprint arXiv:2504.13176 (2025).

\bibitem{shen2025long}
F.~Shen, C.~Wang, J.~Gao, Q.~Guo, J.~Dang, J.~Tang, T.-S. Chua, Long-term talkingface generation via motion-prior conditional diffusion model, arXiv preprint arXiv:2502.09533 (2025).

\bibitem{chen2014semantic}
L.-C. Chen, G.~Papandreou, I.~Kokkinos, K.~Murphy, A.~L. Yuille, Semantic image segmentation with deep convolutional nets and fully connected crfs, arXiv preprint arXiv:1412.7062 (2014).

\bibitem{chen2017deeplab}
L.-C. Chen, G.~Papandreou, I.~Kokkinos, K.~Murphy, A.~L. Yuille, Deeplab: Semantic image segmentation with deep convolutional nets, atrous convolution, and fully connected crfs, IEEE transactions on pattern analysis and machine intelligence 40~(4) (2017) 834--848.

\bibitem{chen2017rethinking}
L.-C. Chen, G.~Papandreou, F.~Schroff, H.~Adam, Rethinking atrous convolution for semantic image segmentation, arXiv preprint arXiv:1706.05587 (2017).

\bibitem{chen2018encoder}
L.-C. Chen, Y.~Zhu, G.~Papandreou, F.~Schroff, H.~Adam, Encoder-decoder with atrous separable convolution for semantic image segmentation, in: Proceedings of the European conference on computer vision (ECCV), 2018, pp. 801--818.

\bibitem{Dai2017DeformableCN}
J.~Dai, H.~Qi, Y.~Xiong, Y.~Li, G.~Zhang, H.~Hu, Y.~Wei, Deformable convolutional networks, in: Proceedings of the IEEE international conference on computer vision, 2017, pp. 764--773.

\bibitem{Howard2017MobileNetsEC}
A.~G. Howard, M.~Zhu, B.~Chen, D.~Kalenichenko, W.~Wang, T.~Weyand, M.~Andreetto, H.~Adam, Mobilenets: Efficient convolutional neural networks for mobile vision applications, ArXiv abs/1704.04861 (2017).

\bibitem{Yang2019CondConvCP}
B.~Yang, G.~Bender, Q.~V. Le, J.~Ngiam, Condconv: Conditionally parameterized convolutions for efficient inference, Advances in neural information processing systems 32 (2019).

\bibitem{Chen2019DynamicCA}
Y.~Chen, X.~Dai, M.~Liu, D.~Chen, L.~Yuan, Z.~Liu, Dynamic convolution: Attention over convolution kernels, 2020 IEEE/CVF Conference on Computer Vision and Pattern Recognition (CVPR) (2019) 11027--11036.

\bibitem{Tan2019MixConvMD}
M.~Tan, Q.~V. Le, Mixconv: Mixed depthwise convolutional kernels, arXiv preprint arXiv:1907.09595 (2019).

\bibitem{Yang2024PinwheelConv}
J.~Yang, S.~Liu, J.~Wu, X.~Su, N.~Hai, X.~Huang, Pinwheel-shaped convolution and scale-based dynamic loss for infrared small target detection, Proceedings of the AAAI Conference on Artificial Intelligence 39~(9) (2025) 9202--9210.

\bibitem{vaswani2017attention}
A.~Vaswani, N.~Shazeer, N.~Parmar, J.~Uszkoreit, L.~Jones, A.~N. Gomez, {\L}.~Kaiser, I.~Polosukhin, Attention is all you need, Advances in neural information processing systems 30 (2017).

\bibitem{Gu2023MambaLS}
A.~Gu, T.~Dao, Mamba: Linear-time sequence modeling with selective state spaces, ArXiv abs/2312.00752 (2023).

\bibitem{Liu2021SwinTH}
Z.~Liu, Y.~Lin, Y.~Cao, H.~Hu, Y.~Wei, Z.~Zhang, S.~Lin, B.~Guo, Swin transformer: Hierarchical vision transformer using shifted windows, in: Proceedings of the IEEE/CVF international conference on computer vision, 2021, pp. 10012--10022.

\bibitem{Liu2024VMambaVS}
Y.~Liu, Y.~Tian, Y.~Zhao, H.~Yu, L.~Xie, Y.~Wang, Q.~Ye, J.~Jiao, Y.~Liu, Vmamba: Visual state space model, Advances in neural information processing systems 37 (2024) 103031--103063.

\bibitem{Zhang2021TransFuseFT}
Y.~Zhang, H.~Liu, Q.~Hu, Transfuse: Fusing transformers and cnns for medical image segmentation, ArXiv abs/2102.08005 (2021).

\bibitem{Xu2024HCMambaVM}
J.~Xu, Hc-mamba: Vision mamba with hybrid convolutional techniques for medical image segmentation, ArXiv abs/2405.05007 (2024).

\bibitem{He2015DeepRL}
K.~He, X.~Zhang, S.~Ren, J.~Sun, Deep residual learning for image recognition, in: Proceedings of the IEEE conference on computer vision and pattern recognition, 2016, pp. 770--778.

\bibitem{dosovitskiy2020image}
A.~Dosovitskiy, L.~Beyer, A.~Kolesnikov, D.~Weissenborn, X.~Zhai, T.~Unterthiner, M.~Dehghani, M.~Minderer, G.~Heigold, S.~Gelly, et~al., An image is worth 16x16 words: Transformers for image recognition at scale, arXiv preprint arXiv:2010.11929 (2020).

\bibitem{Gu2021EfficientlyML}
A.~Gu, K.~Goel, C.~R'e, Efficiently modeling long sequences with structured state spaces, ArXiv abs/2111.00396 (2021).

\bibitem{Gu2022OnTP}
A.~Gu, A.~Gupta, K.~Goel, C.~R{\'e}, On the parameterization and initialization of diagonal state space models, ArXiv abs/2206.11893 (2022).

\bibitem{Zhu2024VisionME}
L.~Zhu, B.~Liao, Q.~Zhang, X.~Wang, W.~Liu, X.~Wang, Vision mamba: Efficient visual representation learning with bidirectional state space model, ArXiv abs/2401.09417 (2024).

\bibitem{Finder2024WaveletCF}
S.~E. Finder, R.~Amoyal, E.~Treister, O.~Freifeld, Wavelet convolutions for large receptive fields, in: European Conference on Computer Vision, Springer, 2024, pp. 363--380.

\bibitem{Deng2009ImageNetAL}
J.~Deng, W.~Dong, R.~Socher, L.-J. Li, K.~Li, L.~Fei-Fei, Imagenet: A large-scale hierarchical image database, in: 2009 IEEE conference on computer vision and pattern recognition, Ieee, 2009, pp. 248--255.

\bibitem{landman2015miccai}
B.~Landman, Z.~Xu, J.~Igelsias, M.~Styner, T.~Langerak, A.~Klein, Miccai multi-atlas labeling beyond the cranial vault--workshop and challenge, in: Proc. MICCAI multi-atlas labeling beyond cranial vault—workshop challenge, Vol.~5, Munich, Germany, 2015, p.~12.

\bibitem{Bernard2018DeepLT}
O.~Bernard, A.~Lalande, C.~Zotti, F.~Cervenansky, X.~Yang, P.-A. Heng, I.~Cetin, K.~Lekadir, O.~Camara, M.~A.~G. Ballester, G.~Sanrom{\'a}, S.~Napel, S.~E. Petersen, G.~Tziritas, E.~Grinias, M.~Khened, V.~A. Kollerathu, G.~Krishnamurthi, M.-M. Roh{\'e}, X.~Pennec, M.~Sermesant, F.~Isensee, P.~F. J{\"a}ger, K.~H. Maier-Hein, P.~M. Full, I.~Wolf, S.~Engelhardt, C.~F. Baumgartner, L.~M. Koch, J.~M. Wolterink, I.~I\v{s}gum, Y.~Jang, Y.~Hong, J.~Patravali, S.~Jain, O.~Humbert, P.-M. Jodoin, Deep learning techniques for automatic mri cardiac multi-structures segmentation and diagnosis: Is the problem solved?, IEEE Transactions on Medical Imaging 37 (2018) 2514--2525.

\bibitem{Rahman2023MedicalIS}
M.~M. Rahman, R.~Marculescu, Medical image segmentation via cascaded attention decoding, in: Proceedings of the IEEE/CVF winter conference on applications of computer vision, 2023, pp. 6222--6231.

\bibitem{xu2023levit}
G.~Xu, X.~Zhang, X.~He, X.~Wu, Levit-unet: Make faster encoders with transformer for medical image segmentation, in: Chinese Conference on Pattern Recognition and Computer Vision (PRCV), Springer, 2023, pp. 42--53.

\bibitem{azad2023dae}
R.~Azad, R.~Arimond, E.~K. Aghdam, A.~Kazerouni, D.~Merhof, Dae-former: Dual attention-guided efficient transformer for medical image segmentation, in: International workshop on predictive intelligence in medicine, Springer, 2023, pp. 83--95.

\bibitem{huang2022scaleformer}
H.~Huang, S.~Xie, L.~Lin, Y.~Iwamoto, X.~Han, Y.-W. Chen, R.~Tong, Scaleformer: revisiting the transformer-based backbones from a scale-wise perspective for medical image segmentation, arXiv preprint arXiv:2207.14552 (2022).

\bibitem{Azad2023BeyondSD}
R.~Azad, L.~Niggemeier, M.~Huttemann, A.~Kazerouni, E.~K. Aghdam, Y.~Velichko, U.~Bagci, D.~Merhof, Beyond self-attention: Deformable large kernel attention for medical image segmentation, 2024 IEEE/CVF Winter Conference on Applications of Computer Vision (WACV) (2023) 1276--1286.

\bibitem{Rahman2023GCASCADEEC}
M.~M. Rahman, R.~Marculescu, G-cascade: Efficient cascaded graph convolutional decoding for 2d medical image segmentation, 2024 IEEE/CVF Winter Conference on Applications of Computer Vision (WACV) (2023) 7713--7722.

\bibitem{Wang2021PVTVI}
W.~Wang, E.~Xie, X.~Li, D.-P. Fan, K.~Song, D.~Liang, T.~Lu, P.~Luo, L.~Shao, Pvt v2: Improved baselines with pyramid vision transformer, Computational visual media 8~(3) (2022) 415--424.

\bibitem{Liu2024SwinUMambaMU}
J.~Liu, H.~Yang, H.-Y. Zhou, Y.~Xi, L.~Yu, C.~Li, Y.~Liang, G.~Shi, Y.~Yu, S.~Zhang, et~al., Swin-umamba: Mamba-based unet with imagenet-based pretraining, in: International Conference on Medical Image Computing and Computer-Assisted Intervention, Springer, 2024, pp. 615--625.

\bibitem{Liu2025MixFormerAM}
J.~Liu, K.~Li, C.~Huang, H.~Dong, Y.~Song, R.~Li, Mixformer: A mixed cnn–transformer backbone for medical image segmentation, IEEE Transactions on Instrumentation and Measurement 74 (2025) 1--20.

\end{thebibliography}

\end{document}